\documentclass{article} 
\usepackage{iclr2027_conference,times}

\usepackage{amsmath,amsfonts,bm}

\def\eqref#1{equation~\ref{#1}}

\def\1{\bm{1}}

\DeclareMathAlphabet{\mathsfit}{\encodingdefault}{\sfdefault}{m}{sl}
\SetMathAlphabet{\mathsfit}{bold}{\encodingdefault}{\sfdefault}{bx}{n}

\usepackage{hyperref}
\usepackage{url}
\usepackage{graphicx}
\usepackage{subcaption}

\usepackage{algorithm}
\usepackage{algpseudocode}

\usepackage{graphicx}
\usepackage{subcaption}
\usepackage{array}
\usepackage{booktabs}
\usepackage{multirow}
\usepackage{tabularx}
\usepackage{makecell}
\usepackage{placeins}
\usepackage{bm}
\newcolumntype{Y}{>{\centering\arraybackslash}X}

\title{
Sequence Composition in EHR Foundation Model Pretraining
}

\author{
\textbf{Joshua Placidi} \qquad
\textbf{Inigo Stibbe} \qquad
\textbf{Alfred J. Balston} \qquad
\textbf{Marek Rei} \qquad
\textbf{A. Aldo Faisal}
\\
Department of Computing, Imperial College London
\\
\texttt{\{joshua.placidi25,marek.rei,a.faisal\}@imperial.ac.uk}
}

\iclrfinalcopy 
\begin{document}

\maketitle
\lhead{Preprint 2026}

\begin{abstract}
Autoregressive electronic health record (EHR) foundation models predict the next event token in patient trajectories: sequences of structured medical events.
A typical pretraining strategy is to concatenate all trajectories into a global stream and sample token windows from it.
This gives patients with longer sequences more training exposure and allows tokens to attend across patient boundaries.
We argue that these choices create a misalignment between pretraining and downstream clinical tasks.
To explore this, we investigate how the composition of pretraining sequences affects clinical prediction performance.
Using the global-stream approach as a baseline, we propose alternative sequence composition methods that control how patient trajectories contribute to pretraining sequences.
Across four MIMIC-IV v2.2 tasks, our best method improves macro AUROC, AUPRC, F1, and balanced accuracy, with relative gains of 9.3\% in F1 and 2.7\% in AUPRC under a fixed training recipe.
We find that these gains generalize across four additional evaluation settings, including the INSPIRE and eICU-CRD datasets, with improvements in macro AUROC, AUPRC, and F1 score.
Our results show that simply changing how EHR data are presented to the model during pretraining can improve downstream performance, without changing the model or collecting additional data.
\end{abstract}

\begin{figure}[!b]
    \centering
    \includegraphics[width=\textwidth]{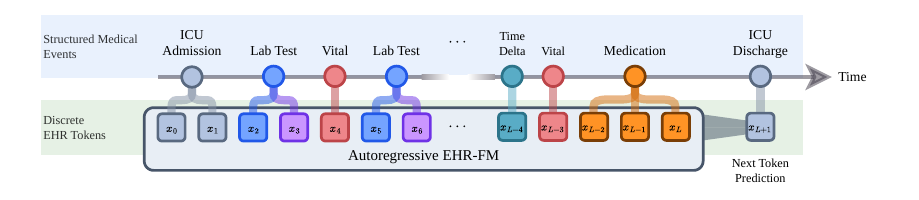}
    \caption{
        Autoregressive prediction from a patient trajectory.
        Medical events are encoded as discrete tokens, and the EHR-FM predicts the next token from the preceding clinical context.
    }
    \label{fig:ehr_sequence}
\end{figure}

\section{Introduction}
Electronic health records (EHRs) represent medical histories as trajectories of events describing diagnoses, procedures, medications, laboratory measurements, and more.
These histories vary widely in length and detail, reflecting differences in patient states and the type of care they require.
EHR foundation models (EHR-FMs) aim to learn representations from these longitudinal records to support a range of clinical tasks.
Autoregressive EHR-FMs represent each trajectory as a sequence of tokens and learn through next-token prediction, as shown in Figure~\ref{fig:ehr_sequence}.
Architectures and training pipelines for these models are often adapted from language modeling frameworks, with medical event tokens replacing text tokens \citep{renc_2024_zero, kraljevic_2024_foresight, fallahpour_2025_ehrmamba}.
Models are evaluated in a zero-shot setting by prompting them with a patient history and stochastically generating future \emph{rollouts}.
The fraction of rollouts containing an outcome provides an estimate of that patient's risk of that outcome \citep{renc_2025_foundation}.
Pretraining therefore directly shapes how these models perform on downstream clinical tasks.

\begin{figure}[t]
\centering
\makebox[\linewidth][c]{%
    \begin{subfigure}[t]{0.25\linewidth}
        \centering
        \includegraphics[width=\linewidth]
            {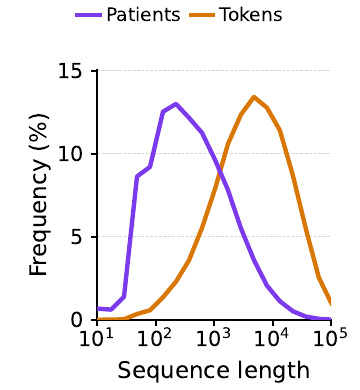}
        \caption{Patients and tokens.}
        \label{fig:patient_vs_token_dist}
    \end{subfigure}%
    \hfill
    \begin{subfigure}[t]{0.25\linewidth}
        \centering
        \includegraphics[width=\linewidth]
            {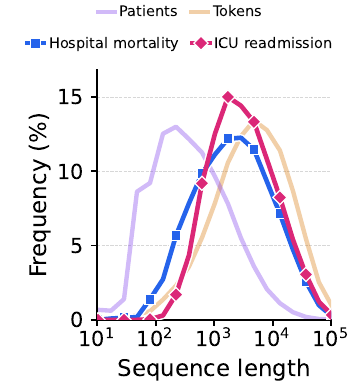}
        \caption{Clinical task instances.}
        \label{fig:clinical_task_dist}
    \end{subfigure}%
    \hfill
    \begin{subfigure}[t]{0.25\linewidth}
        \centering
        \includegraphics[width=\linewidth]
            {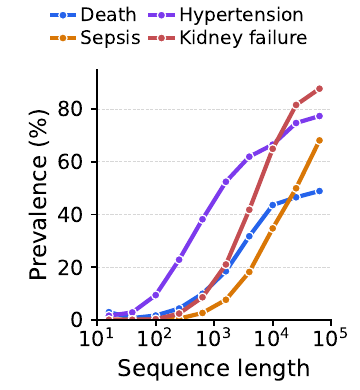}
        \caption{Outcome prevalence.}
        \label{fig:outcome_prevalence}
    \end{subfigure}%
    \hfill
    \begin{subfigure}[t]{0.25\linewidth}
        \centering
        \includegraphics[width=\linewidth]
            {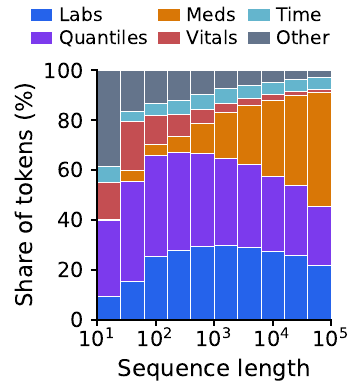}
        \caption{Token composition.}
        \label{fig:trajectory_composition}
    \end{subfigure}%
}

\caption{
Trajectory token length and clinical content in the training set of MIMIC-IV v2.2.
(a) Distribution of patient trajectory lengths and tokens. Each token is represented by the length of its corresponding trajectory.
(b) Distributions of hospital mortality and ICU readmission task instances by trajectory length.
(c) Recorded outcome prevalence by trajectory length.
(d) Token composition by trajectory length.
}

\label{fig:trajectory_distribution}
\end{figure}

The method of composing pretraining sequences from a dataset of patient trajectories determines how each patient contributes to model training.
This choice is particularly consequential for zero-shot EHR-FMs, because they do not undergo additional task fine-tuning.
A common composition method is to concatenate patient trajectories into a global token stream and sample fixed-length windows from that combined stream \citep{renc_2024_zero,renc_2025_foundation}.
These windows can contain information from multiple patients packed together without any boundary masking, allowing tokens from one patient to attend to tokens from another.
Allowing attention across unrelated documents can introduce distracting context and degrade downstream performance in language-model pretraining \citep{zhao_2024_analysing}. In EHR pretraining, this corresponds to conditioning on events from unrelated patients, despite downstream predictions being made from individual patient histories.

Using MIMIC-IV v2.2 \citep{PhysioNet-mimiciv-2.2, mimiciv-paper, PhysioNet-mimic-iv-ed-2.2}, we observe that patient trajectory lengths span several orders of magnitude due to differences in duration and severity of condition.
Sampling global-stream windows gives longer histories proportionally more next-token targets, so exposure follows the token rather than the patient distribution.
Figure~\ref{fig:patient_vs_token_dist} shows that token exposure is shifted toward much longer trajectories than the patient distribution, peaking near $10^4$ versus $10^2$ tokens.
Mortality prediction at hospital admission and readmission prediction at ICU discharge are common EHR-FM benchmarks.
Figure~\ref{fig:clinical_task_dist} shows that instances for both tasks peak at trajectory lengths of approximately $10^3$ tokens.
Clinical content also varies with length.
Recorded outcome prevalence rises in longer trajectories (Figure~\ref{fig:outcome_prevalence}), while the mix of token types changes (Figure~\ref{fig:trajectory_composition}).
Reweighting trajectories therefore changes both patient exposure and the clinical information encountered during pretraining.

\begin{figure*}[t]
    \centering
    \includegraphics[
        width=\textwidth
    ]{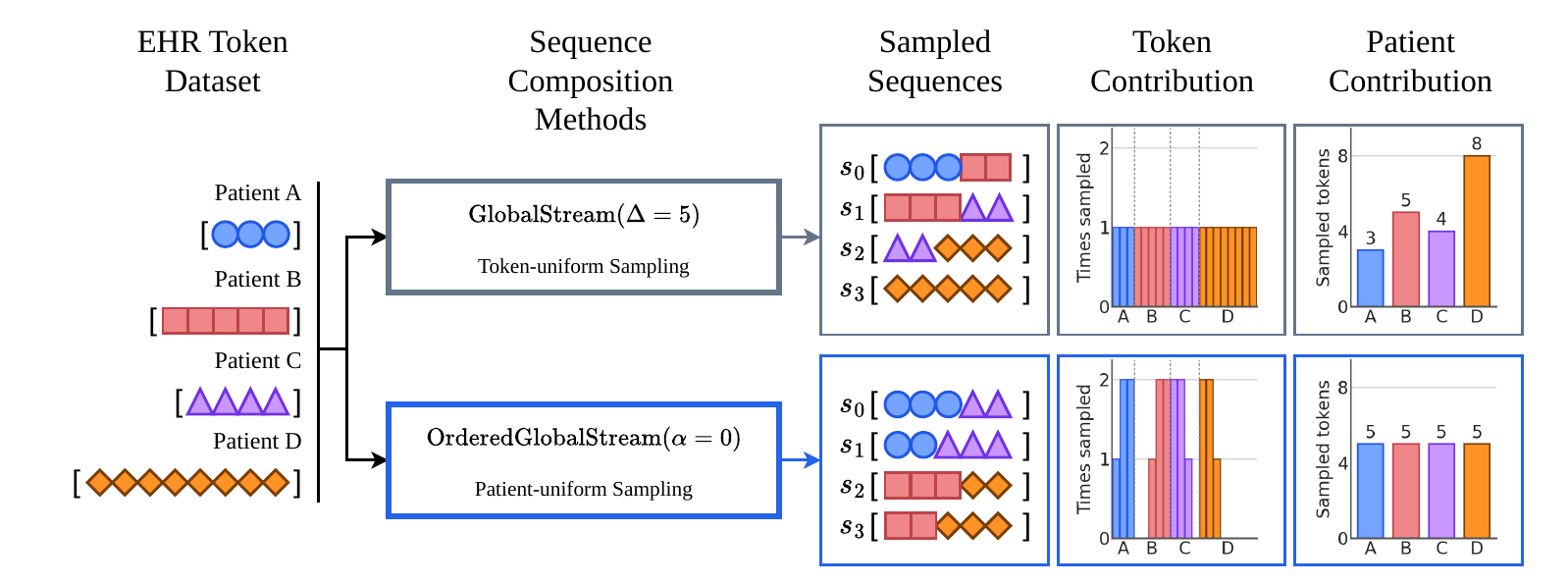}

    \caption{
Sequence composition for four patients with unequal trajectory lengths.
In the Global Stream example, each token appears once, so longer trajectories contribute more tokens.
Ordered Global Stream ($\alpha=0$) samples windows repeatedly to give patients approximately equal expected contributions.
    }
    \label{fig:abstract}
\end{figure*}

We study sequence composition methods under a fixed model architecture and training recipe.
We compare Global Stream (GS) with a masked version (GS+M) that prevents attention crossing patient boundaries.
We then introduce three methods for controlling patient contribution: Patient Sampling (PS), Recursive Patient Sampling (RPS), and Ordered Global Stream (OGS).
Figure~\ref{fig:abstract} illustrates how composition methods can generate different sequences, modifying the token and patient contributions during training.
This comparison allows us to examine the impact of cross-patient attention and sequence composition on zero-shot clinical prediction.

Across four MIMIC-IV v2.2 tasks, OGS achieves the highest mean performance on macro AUROC, AUPRC, F1, and balanced accuracy across five seeds.
Relative to GS, OGS improves macro AUPRC by 2.7\%, F1 by 9.3\%, and balanced accuracy by 3.7\%.
We further evaluate OGS in four additional settings, including the INSPIRE and eICU-CRD datasets.
Relative to GS+M, OGS improves mean macro AUROC, AUPRC, and F1 across these settings by 2.0\%, 2.3\%, and 8.2\%, respectively, with gains in all three metrics in every setting.
These findings establish pretraining sequence composition as an important design choice for EHR foundation models.

\section{Related Work}

\subsection{Autoregressive EHR-FMs}

Prior work has primarily focused on tokenizing EHRs for language-model-style pretraining and developing downstream evaluation frameworks for EHR-FMs.
CLMBR demonstrates that next-event prediction can yield transferable patient representations, particularly when labelled data are limited \citep{steinberg_2021_language}.
EHRSHOT subsequently provides a benchmark for evaluating pretrained EHR models across clinical prediction tasks \citep{wornow_2023_ehrshot}.
However, the CLMBR-T model evaluated with EHRSHOT is pretrained on a separate 2.57M-patient Stanford cohort, which is not publicly released, limiting direct reproduction and comparison.
Foresight and ETHOS extend next-event prediction to generative forecasting of patient trajectories \citep{kraljevic_2024_foresight,renc_2024_zero}.
ETHOS proposes a tokenization framework that converts MIMIC-IV \citep{mimiciv-paper} records into discrete token trajectories \citep{renc_2024_zero}.
The authors train ETHOS using next-token prediction and evaluate it using a rollout-based zero-shot clinical prediction framework.
They show that autoregressive EHR-FMs can achieve competitive performance against supervised baselines without task-specific training.
More recent work has studied long-context architectures, context length, and scaling with compute, model size, and data volume \citep{fallahpour_2025_ehrmamba,wornow_2025_context,zhang_2025_exploring,waxler_2025_generative}, while comparatively little attention has been paid to how patient trajectories are composed into pretraining sequences.

\subsection{Sequence Composition and Document Packing}

Language-model pretraining often packs documents into fixed-length sequences to use the available context efficiently \citep{ding_2024_fewer}.
Packing determines both which documents appear together and which information tokens can attend to.
With standard causal attention, a token can attend to earlier tokens from an unrelated document in the same sequence.
\citet{zhao_2024_analysing} study this cross-document attention and show that restricting attention to tokens within each document can improve downstream behaviour.
Changing the contents of a sequence offers another way to shape context. \citet{shi_2024_incontext} find that grouping semantically related documents improves performance relative to random packing.
These results distinguish two decisions that can also arise with patient trajectories: how histories are placed into sequences, and whether attention crosses their boundaries.

\subsection{Sampling and Reweighting}

Sampling and reweighting change how often different parts of a dataset contribute to the training objective.
DoReMi learns a mixture over language-model pretraining domains using a proxy model, without selecting weights against particular downstream tasks \citep{xie_2023_doremi}.
They show that training exposure can affect downstream performance even when the underlying data sources remain the same.
Clinical prediction work has likewise used patient reweighting to address temporal and cross-institutional shifts, and downsampling or sample weighting to address rare outcomes \citep{lee_2023_stable,zhu_2025_fedweight,yuan_2022_autoscoreimbalance}.
In autoregressive EHR pretraining, sampling determines how frequently each patient's tokens provide prediction targets.
Taken together, this work motivates studying patient exposure and training context as linked consequences of sequence composition.

\begin{algorithm}[t]
\caption{Global Stream (GS)}
\label{alg:global-stream}
\begin{algorithmic}[1]
\Require Ordered patient timelines
$\mathcal{D}=(\mathbf{x}^{(1)},\ldots,\mathbf{x}^{(N)})$,
context length $L$, stride $\Delta$
\State $\mathbf{g} \gets
\mathbf{x}^{(1)} \Vert \cdots \Vert \mathbf{x}^{(N)}$
\Comment{Concatenate trajectories}
\State $K \gets 1+\left\lfloor
\frac{|\mathbf{g}|-L-1}{\Delta}
\right\rfloor$
\Comment{Number of valid windows}
\State $k \sim \operatorname{Uniform}\{0,\ldots,K-1\}$
\Comment{Sample a window index}
\State $j \gets 1+k\Delta$
\State $\mathbf{s} \gets (g_j,\ldots,g_{j+L})$
\Comment{Extract an $(L+1)$-token window}
\State \Return $\mathbf{s}$
\end{algorithmic}
\end{algorithm}

\begin{algorithm}[t]
\caption{Patient Sampling (PS)}
\label{alg:patient-sampling}
\begin{algorithmic}[1]
\Require Patient timelines
$\mathcal{D}=(\mathbf{x}^{(1)},\ldots,\mathbf{x}^{(N)})$,
context length $L$, patient sampling distribution $p_\alpha$
\State $i \sim \operatorname{Categorical}(p_\alpha)$
\Comment{Weighted patient sampling}
\State $T_i \gets |\mathbf{x}^{(i)}|$
\State $j \sim \operatorname{Uniform}\{2-L,\ldots,T_i-1\}$
\Comment{Sample a boundary-overlapping window}
\State $\mathbf{s} \gets
\left(x_{\max(1,j)}^{(i)},\ldots,
x_{\min(T_i,j+L)}^{(i)}\right)$
\Comment{Clip to the patient trajectory}
\State $\mathbf{s} \gets \Call{PadRight}{\mathbf{s},L+1}$
\State \Return $\mathbf{s}$
\end{algorithmic}
\end{algorithm}

\section{Sequence Composition Methods}

We pretrain on a tokenized EHR dataset $\mathcal{D}=\{\mathbf{x}^{(1)},\ldots,\mathbf{x}^{(N)}\}$ containing $N$ patients.
Here, $\mathbf{x}^{(i)}=(x_1^{(i)},\ldots,x_{T_i}^{(i)})$ is the chronologically ordered token sequence for patient $i$, with length $T_i$.
Given a context length $L$, each sequence-composition method constructs a training sequence $\mathbf{s}=(s_1,\ldots,s_{L+1})$ from $\mathcal{D}$, padding any unused positions.
The model receives $\mathbf{s}_{1:L}$ as input and predicts the shifted targets $\mathbf{s}_{2:L+1}$.

\subsection{Global Stream}

We use Global Stream (GS) as our baseline, following the concatenated-timeline pretraining approach used by ETHOS \citep{renc_2024_zero}. GS concatenates patient trajectories into a token stream $\mathbf{g}$ and defines candidate windows of $L+1$ tokens at starting positions separated by stride $\Delta$ (Algorithm~\ref{alg:global-stream}).
During training, these windows are sampled uniformly with replacement.
Longer trajectories occupy more positions in $\mathbf{g}$ and therefore contribute approximately proportionally more next-token training targets.
Windows may span patient boundaries, allowing tokens to attend to preceding patients under standard causal attention.
To assess the effect of cross-patient attention, we also evaluate Global Stream with masking (GS+M).
It samples the same windows as GS but prevents attention across patient boundaries.
The methods we introduce below also prevent cross-patient attention, so GS+M lets us assess the effects of sequence composition separately from those of attention masking.

\subsection{Patient Contribution Weighting}
\label{sec:patient-weighting}

In our subsequent methods, we control the contribution of each patient to pretraining through their share of next-token prediction targets.
Let $M_i$ be the number of loss-eligible next-token targets in patient $i$'s trajectory.
For $\alpha\in[0,1]$, we define the target share as
\begin{equation}
\pi_i(\alpha)
=
\frac{M_i^\alpha}{\sum_{j=1}^{N} M_j^\alpha}
=
\operatorname{softmax}_i(\alpha\log M_i).
\label{eq:target-patient-contribution}
\end{equation}
At $\alpha=0$, every patient has an equal target share; at $\alpha=1$, shares are proportional to $M_i$, approximately matching the token-proportional exposure of GS.
Intermediate values shift exposure between these endpoints, as illustrated for Patient Sampling in Appendix Figure~\ref{fig:loss-share-vs-alpha}.
The methods below use different sampling procedures to approach these target shares.

\subsection{Patient Sampling}
\label{sec:patient-sampling}
GS treats the dataset as a stream of tokens.
Patient Sampling (PS) is a simple method that maintains the patient-centric structure of the dataset.
PS first selects a single patient, then samples a contiguous window from that patient's trajectory (Algorithm~\ref{alg:patient-sampling}).
PS never combines tokens from different patients in the same training sequence.
Short patient trajectories are right-padded to length $L+1$, with padding tokens excluded from the loss.
To preserve the desired training exposure, the patient sampling weights must therefore account for the number of valid prediction targets contributed by each trajectory.
Let $E_i$ be the expected number of targets in a window sampled from patient $i$.
We account for this variation by drawing patient $i$ with probability $p_\alpha(i)\propto\pi_i(\alpha)/E_i$, aiming to match the desired target shares.
The normalized probabilities and their derivation are given in Appendix~\ref{app:patient-contribution-weighting}.

\subsection{Recursive Patient Sampling}
\label{sec:recursive-patient-sampling}

\begin{algorithm}[t]
\caption{Recursive Patient Sampling (RPS)}
\label{alg:recursive-patient-sampling}
\begin{algorithmic}[1]
\Require Patient timelines $\mathcal{D}$, context length $L$,
patient sampling distribution $p_\alpha$
\State $\mathbf{s}\gets()$
\State $R\gets L+1$ \Comment{Remaining capacity}
\While{$R\geq 2$}
    \State $i\sim\operatorname{Categorical}(p_\alpha)$
    \State $T_i \gets |\mathbf{x}^{(i)}|$
    \State $j\sim\operatorname{Uniform}\{3-R,\ldots,T_i-1\}$
        \Comment{Sample a boundary-overlapping window}
    \State $\mathbf{c}\gets
        \left(x_{\max(1,j)}^{(i)},\ldots,
        x_{\min(T_i,j+R-1)}^{(i)}\right)$
        \Comment{Clip to the patient trajectory}
    \State $\mathbf{s}\gets\mathbf{s}\Vert\mathbf{c}$
    \State $R\gets R-|\mathbf{c}|$
    \Comment{Update the remaining capacity in the sequence}
\EndWhile
\State $\mathbf{s}\gets\Call{PadRight}{\mathbf{s},L+1}$
\State \Return $\mathbf{s}$
\end{algorithmic}
\end{algorithm}


\begin{algorithm}[t]
\caption{Ordered Global Stream (OGS)}
\label{alg:ordered-global-stream}
\begin{algorithmic}[1]
\Require Patient timelines $\mathcal{D}$, context length $L$,
stride $\Delta$, exponent $\alpha$
\State $\boldsymbol{\rho}\gets\operatorname{argsort}_i(T_i)$
    \Comment{Order patients by length}
\State $\mathbf{g}\gets
    \mathbf{x}^{(\rho_1)}\Vert\cdots\Vert\mathbf{x}^{(\rho_N)}$
    \Comment{Concatenate in this order}
\State $\mathcal{I}\gets
    \{1+k\Delta:k\geq0,\ 1+k\Delta+L\leq|\mathbf{g}|\}$
    \Comment{The set of valid window starting indices}
\State $q_n\gets M_{\text{pat}(n)}^\alpha/T_{\text{pat}(n)}
    \quad\text{for }n=1,\ldots,|\mathbf{g}|$
    \Comment{Distribute patient weight across tokens}
\Statex \hspace{\algorithmicindent}
    \textit{Where $\text{pat}(n)$ is the patient at position $n$ in $\mathbf{g}$.}
\State $w_j\gets\frac{1}{L+1}\sum_{n=j}^{j+L}q_n
    \quad\text{for }j\in\mathcal{I}$
    \Comment{Calculate mean of token weights in each window}
\State $p_j\gets w_j\Big/\sum_{r\in\mathcal I}w_r
    \quad\text{for }j\in\mathcal I$
    \Comment{Normalize window weights}
\State $j\sim\operatorname{Categorical}(\{p_r\}_{r\in\mathcal I})$
    \Comment{Sample a window starting index}
\State \Return $(g_j,\ldots,g_{j+L})$
\end{algorithmic}
\end{algorithm}

Patient Sampling can leave much of a sequence padded when a patient's trajectory is short.
Recursive Patient Sampling (RPS) fills the unused space with additional patient chunks (Algorithm~\ref{alg:recursive-patient-sampling}).
It repeatedly draws a patient, samples a contiguous chunk that fits the remaining capacity, and appends it to the sequence.
The process stops when fewer than two positions remain, and any unused position is padded.
Chunks may come from different patients or from noncontiguous parts of the same trajectory.
We mask the next-token target at each join so the model is not trained to predict across unrelated chunks.
We also mask attention across all chunk boundaries.
RPS aims for the same target shares $\pi_i(\alpha)$ as PS, but its patient-draw weights differ because the capacity available for each chunk changes during packing.
The weighting procedure is described in Appendix~\ref{app:recursive-patient-sampling-weights}.

\subsection{Ordered Global Stream}
\label{sec:ordered-global-stream}

Ordered Global Stream (OGS) aims to control patient contributions while retaining the efficient structure of GS, without the need for padding tokens or recursive packing (Algorithm~\ref{alg:ordered-global-stream}).
We sort patients by trajectory length before concatenating their timelines, placing patients of similar length near one another.
We assign each token belonging to patient $i$ a weight of $M_i^\alpha/T_i$, distributing that patient's weight across their trajectory.
Windows that cross patient boundaries contain tokens with similar sampling weights due to the sorted order of trajectories.
The weights of patient $i$'s tokens sum to $M_i^\alpha$.
Each candidate window receives the mean weight of the $L+1$ tokens it contains, and we sample windows in proportion to these weights.
At $\alpha=0$, patients have equal total token weight; at $\alpha=1$, token weights are approximately uniform for longer trajectories, making window sampling the same as GS.
Because weights are assigned to windows rather than individual prediction targets, the resulting patient contributions approximate $\pi_i(\alpha)$.
Appendix~\ref{app:ordered-global-stream-weights} provides the weighting details.

\section{Experimental Setup}

\begin{table*}[t]
\centering
\small
\setlength{\tabcolsep}{3pt}
\caption{Dataset and tokenization statistics. Token counts and trajectory
lengths are calculated after tokenization. The final column gives the
percentage of patients whose complete trajectory exceeds 2,048 tokens, which is the size of the context window, $L$, we use in our experiments.}
\label{tab:datasets}
\begin{tabular*}{\textwidth}{@{\extracolsep{\fill}}llrrrrr@{}}
\toprule
Dataset & Tokenization
& \multicolumn{1}{c}{Patients}
& \multicolumn{1}{c}{Tokens}
& \multicolumn{1}{c}{\makecell{Vocab.\\size}}
& \multicolumn{1}{c}{\makecell{Median length\\(tokens)}}
& \multicolumn{1}{c}{\makecell{Length $>2{,}048$\\(\% of patients)}} \\
\midrule
MIMIC-IV v2.2 & ETHOS & 285,606 & 362,697,213 & 4,509  & 338   & 14.2\% \\
MIMIC-IV v3.1 & ETHOS & 346,262 & 500,700,081 & 4,598  & 365   & 16.1\% \\
\midrule
MIMIC-IV v2.2 & Ours  & 287,511 & 576,657,770 & 6,431  & 512   & 22.2\% \\
INSPIRE        & Ours  & 99,900  & 176,560,774 & 1,313  & 840   & 18.3\% \\
eICU-CRD       & Ours  & 138,963 & 461,065,895 & 8,000 & 1,719 & 43.1\% \\
\bottomrule
\end{tabular*}
\end{table*}

\subsection{Datasets}
\label{sec:datasets}

Our primary experiments use MIMIC-IV v2.2 \citep{PhysioNet-mimiciv-2.2,mimiciv-paper,PhysioNet-mimic-iv-ed-2.2}, a widely used EHR dataset, to compare sequence-composition methods and patient-weighting
settings.
Our MIMIC-IV data include hospital and intensive care unit (ICU) records from Beth Israel Deaconess Medical Center in Boston, together with linked emergency department (ED) records.
We assess generalization using three additional datasets.
MIMIC-IV v3.1 \citep{PhysioNet-mimiciv-3.1,mimiciv-paper} is a later release of the same data with additional patient records.
eICU-CRD v2.0 \citep{PhysioNet-eicu-crd-2.0,eicu-paper} contains ICU records from multiple hospitals across the United States.
INSPIRE v1.4.2 \citep{PhysioNet-inspire-1.4.2,inspire-paper} contains perioperative records from Seoul National University Hospital in South Korea.
All datasets were obtained through PhysioNet \citep{physio-net-paper}.
We split each dataset by patient into 80\% training, 10\% validation, and 10\% test sets.
We train a new model from scratch for each experiment using the corresponding training split.
Table~\ref{tab:datasets} summarizes each of the datasets used.

\subsection{Tokenization}

\paragraph{ETHOS tokenization.}
For the primary MIMIC-IV v2.2 experiments, we tokenize patient records using the code released with ETHOS \citep{renc_2025_foundation}. We apply the same preprocessing and tokenization configuration to MIMIC-IV v3.1. For both releases, we include the hospital (\texttt{hosp}), intensive care unit (\texttt{icu}), and emergency department (\texttt{ed}) modules. The resulting trajectories encode demographics, clinical events, laboratory values, and elapsed time.

\paragraph{Unified EHR tokenization.}
We develop a general tokenization pipeline that can be applied to MIMIC-IV v2.2, eICU-CRD, and INSPIRE, converting source records to Medical Event Data Standard (MEDS) events \citep{MEDS} through dataset-specific adapters and applying a shared tokenization configuration.
Clinical codes, measurements, units, and elapsed time are represented as discrete tokens, with numeric measurements assigned to 10 quantile bins.
An algorithm inspired by Byte-Pair Encoding \citep{BPE} merges frequent adjacent tokens within the same medical event.
For both pipelines, we fit the tokenizer separately on each dataset's training split and apply it to the corresponding validation and test splits.
Appendix~\ref{app:tokenization} details the encoding and merging procedures and compares the two pipelines.

\begin{figure*}[t]
    \centering
    \includegraphics[
        width=\textwidth
    ]{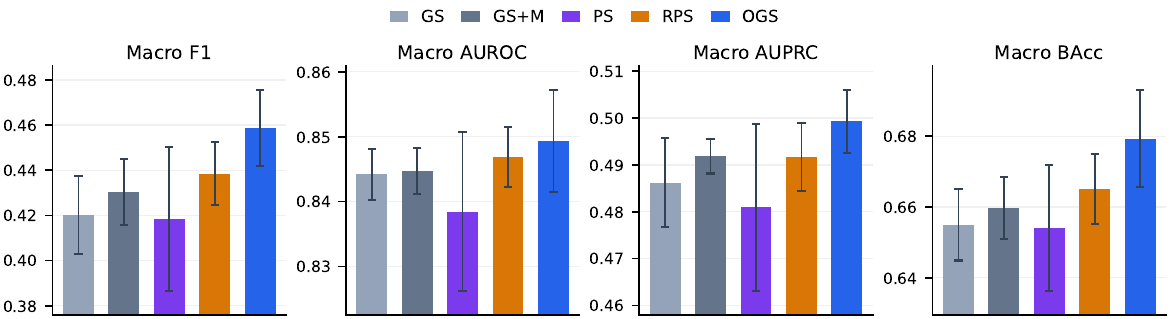}

    \caption{
Macro downstream performance across four clinical prediction tasks on
MIMIC-IV v2.2 test set, reported as mean $\pm$ standard deviation over five seeds.
    }
    \label{fig:e4_summary_bars}
\end{figure*}

\subsection{Model and Training Recipe}
\label{sec:training_recipe}

All experiments use a GPT-2-style decoder-only Transformer \citep{GPT2} with six layers, a hidden dimension of 768, 12 attention heads, and a context length of $L=2048$. Each model has approximately 50M parameters, with the exact count varying by vocabulary size.
The loss is averaged over all loss-eligible targets in each batch. We hold the number of optimisation steps, batch size, and context length fixed across methods; the number of loss-eligible targets varies with sequence composition.
Appendix Figure~\ref{fig:training_data_coverage_vs_alpha} compares expected training-data coverage under a fixed window budget.
Padding and demographic tokens are excluded from the loss. Because each patient sequence or chunk begins with demographic tokens, prediction targets at their boundaries are also excluded.
For RPS, we also mask targets at chunk joins.
GS uses standard causal attention. GS+M, RPS, and OGS prevent attention across patient boundaries using the FlashAttention masking implementation of \citet{zhao_2024_analysing}.

We train each model for 500,000 optimization steps on a single NVIDIA H100 GPU. We select the checkpoint with the lowest next-token validation loss. Validation uses global-stream windows from the corresponding validation split, with standard causal attention for GS and patient-boundary masking for all other methods. For the primary comparison, we use the same five random seeds for each method and report the mean and standard deviation.
Table~\ref{tab:training-recipe} describes the training recipe, with further details in Appendix~\ref{app:training_recipe}.

\subsection{Evaluation Protocol}
\label{sec:evaluation}

\paragraph{Zero-shot clinical prediction.}
We evaluate pretrained models without task-specific fine-tuning. For each task, we prompt the model with a patient's trajectory up to the prediction time and generate 20 independent rollouts with a temperature of 1.
Generation stops when a task-specific stopping condition is met or the maximum number of steps is reached.
Rollouts that reach the step limit without producing a valid terminating token are disregarded.
We estimate risk as the fraction of completed rollouts that reach the positive outcome.

\paragraph{Clinical tasks.}
The primary MIMIC-IV v2.2 evaluation covers hospital mortality, ICU mortality, ICU readmission, and ICU admission.
The mortality tasks predict death during the corresponding stay.
ICU readmission predicts a further ICU stay after ICU discharge, while ICU admission predicts entry to the ICU after hospital admission.
For the other datasets, we evaluate tasks supported by their recorded events.
We report F1, AUROC, AUPRC, and balanced accuracy (BAcc) for each task, using a risk threshold of 0.5 for F1 and BAcc.
Macro scores are unweighted means across the tasks in each dataset and tokenization setting.
Appendix~\ref{app:evaluation} specifies the full evaluation protocol and task definitions across all datasets.

\begin{figure*}[t]
    \centering
    \includegraphics[
        width=\textwidth
    ]{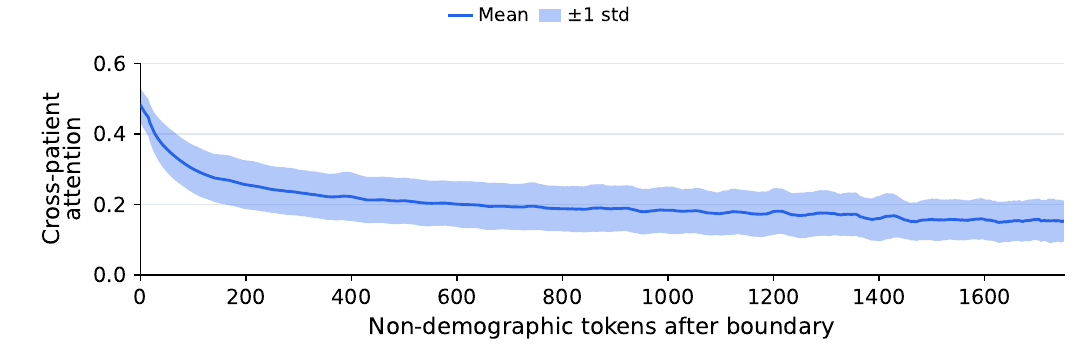}

    \caption{
Cross-patient attention following a patient boundary in GS.
Cross-patient attention is the fraction of attention assigned to tokens belonging to preceding patients.
The line shows the mean across examples, with shading indicating $\pm 1$ standard deviation.
    }
    \label{fig:cross-patient-attention}
\end{figure*}

\begin{figure*}[t]
    \centering
    \includegraphics[
        width=\textwidth
    ]{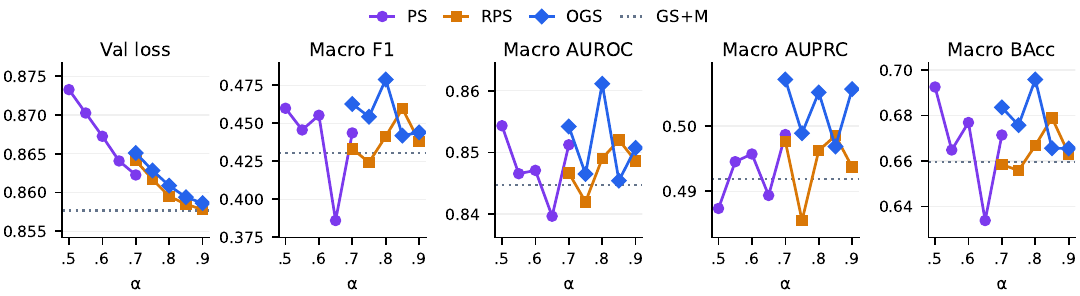}

    \caption{
Local patient-weighting sweeps on MIMIC-IV v2.2. Panels show next-token validation loss and macro validation performance across four clinical tasks as $\alpha$ varies. Dotted lines indicate the GS+M baseline.
    }
    \label{fig:e3_summary_vs_alpha}
\end{figure*}

\begin{figure*}[t]
    \centering
    \includegraphics[
        width=\textwidth
    ]{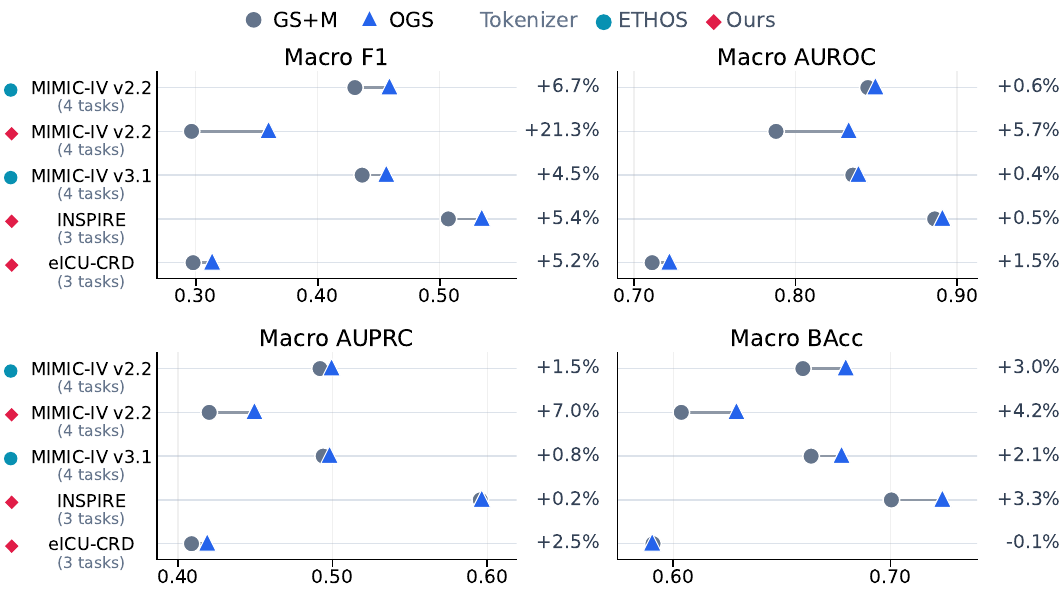}

    \caption{
Generalization across datasets and tokenizations.
Points show macro test-set performance across the indicated tasks, averaged over five seeds for MIMIC-IV v2.2 with ETHOS tokenization and reported for one seed in each additional setting.
Annotations report relative changes from GS+M to OGS.
Markers beside dataset labels identify tokenizers.
Horizontal scales differ between metrics.
    }
    \label{fig:generalization}
\end{figure*}

\section{Results and Analysis}

\subsection{Sequence Composition on MIMIC-IV}
\label{sec:mimic-results}

We compare PS, RPS, and OGS with the GS and GS+M baselines on MIMIC-IV v2.2 using ETHOS tokenization.
We select $\alpha$ separately for each proposed method through coarse and local sweeps, maximizing the mean of macro AUROC and macro AUPRC on the validation set.
This yields $\alpha=0.7$ for PS, $0.85$ for RPS, and $0.8$ for OGS.
Figure~\ref{fig:e4_summary_bars} reports zero-shot performance across four clinical tasks as the mean and standard deviation over five seeds.
Masking attention across patient boundaries in GS produces small improvements across all four macro metrics.
Figure~\ref{fig:cross-patient-attention} shows that tokens immediately following a patient boundary assign substantial attention to preceding patients when this mask is absent. These tokens belong to unrelated clinical histories, yet the GS model uses them as context.
Among the proposed methods, RPS improves most macro metrics over the baselines, whereas PS has lower mean performance and greater variation across seeds.
OGS achieves the highest mean on all four metrics, including when compared with GS+M, which uses the same patient-boundary attention mask.
Relative to GS, OGS improves macro F1, AUROC, AUPRC, and balanced accuracy by 9.3\%, 0.6\%, 2.7\%, and 3.7\%, respectively. 

\subsection{Patient Weighting and Validation Loss}

Figure~\ref{fig:e3_summary_vs_alpha} shows performance near the selected $\alpha$ values; Tables~\ref{tab:initial-alpha-sweep} and~\ref{tab:local-alpha-search} report the full initial and local sweeps.
Next-token validation loss decreases as $\alpha$ increases, reaching its minimum at $\alpha=1$ for all three methods.
However, macro AUROC and AUPRC vary across the local sweep rather than improving steadily with validation loss.
For OGS, $\alpha=0.8$ yields macro AUROC and AUPRC of 0.861 and 0.505, compared with 0.834 and 0.481 at $\alpha=1$, despite the latter's lower validation loss.
These results show that lower next-token validation loss does not necessarily translate into better downstream clinical performance.

\subsection{Generalization Across Datasets and Tokenizations}

We compare OGS ($\alpha=0.8$) with GS+M in four additional settings: MIMIC-IV v2.2 with our unified tokenizer, MIMIC-IV v3.1 with ETHOS tokenization, and INSPIRE and eICU-CRD with our unified tokenizer.
In each case, we train both models from scratch with the same configuration and patient-boundary mask.
Figure~\ref{fig:generalization} shows the four additional experiments alongside the primary MIMIC-IV v2.2 result; Table~\ref{tab:generalization} reports task-level scores. Due to computational cost, we use one seed per method in each additional experiment.
Across the four additional settings, OGS improves averaged macro F1, AUROC, and AUPRC by 8.2\%, 2.0\%, and 2.3\%, respectively.
All three of these metrics improve across all four generalization experiments; balanced accuracy improves in three of four.
The gains vary by task: on eICU-CRD, ICU-readmission F1 falls from 0.411 to 0.343 despite gains on the other two tasks.
Full generalization results are provided in Table~\ref{tab:generalization}.

\section{Conclusion}

Pretraining sequence composition shapes patient contributions to learning and the clinical context available for prediction.
Across four MIMIC-IV v2.2 tasks, OGS achieved the highest mean performance on all four macro metrics, including relative improvements of 9.3\% in macro F1, 2.7\% in AUPRC, and 3.7\% in balanced accuracy over GS.
The improvement over GS+M shows that sequence composition improves clinical prediction even when attention is already restricted to individual patients.
Using $\alpha=0.8$ selected on MIMIC-IV v2.2, we evaluated OGS across additional datasets and tokenization pipelines without retuning patient weighting.
OGS improved macro AUROC, AUPRC, and F1 over GS+M in all four additional experiments, with relative gains in average macro scores of 2.0\%, 2.3\%, and 8.2\%, respectively.
Evaluation was limited to one architecture and training recipe; future work should assess whether these gains persist across model configurations and other clinical settings.
These results support sequence composition as a practical design choice for EHR foundation models: reorganizing existing patient records can improve zero-shot clinical prediction without changing the model architecture.

\pagebreak

\subsection*{Acknowledgments}
JP, IS, and AJB are supported by the UKRI AI Centre for Doctoral Training in Digital Healthcare grant number (EP/Y030974/1).
JP is supported by a philanthropic Bupa PhD Scholarship.
AJB is supported by The Lee Family Faculty of Medicine Scholarship.
AAF and MR acknowledge support from the UKRI AI programme and the Engineering and Physical Sciences Research Council (EPSRC) for the AI Hub in Generative Models (EP/Y028805/1).
AAF \& MR acknowledge the use of resources provided by the Isambard-AI National AI Research Resource (AIRR).
Isambard-AI is operated by the University of Bristol and is funded by the UK Government’s Department for Science, Innovation and Technology (DSIT) via UK Research and Innovation, and the Science and Technology Facilities Council (ST/AIRR/I-A-I/1023).

\pagebreak

\bibliography{iclr2027_conference}
\bibliographystyle{iclr2027_conference}

\pagebreak
\appendix

\FloatBarrier
\section{Patient Contribution Weights}
\label{app:patient-contribution-weighting}

Let $T_i$ be the length of patient $i$'s tokenized trajectory, and let
$M_i$ be its number of loss-eligible next-token targets.
In our implementation, we exclude demographic tokens (such as age and gender) from our loss calculations following \citet{renc_2025_foundation}. Padding tokens are also excluded from the loss.
The weights aim to give patient $i$ a share of training targets
proportional to $M_i^\alpha$.
The counts below describe sampled tokens
before demographic prefixes are added to the sequence.

\subsection{Patient Sampling}
\label{app:patient-sampling-weights}

PS draws a window start uniformly from
$\{2-L,\ldots,T_i-1\}$, giving $A_i=T_i+L-2$ possible starts.
Before demographic processing, each eligible target occurs in exactly
$L$ of these windows. The expected number of targets from a draw of
patient $i$ is therefore
\begin{equation}
E_i=\frac{L M_i}{T_i+L-2}.
\label{eq:expected-targets-per-window}
\end{equation}
We divide the desired target weight by this expected count to obtain
\begin{equation}
p_\alpha(i)
=
\frac{M_i^\alpha/E_i}{\sum_j M_j^\alpha/E_j}
=
\operatorname{softmax}_i\!\left(\alpha\log M_i-\log E_i\right).
\label{eq:patient-sampling}
\end{equation}
Thus, before demographic processing, the expected target share is
\begin{equation}
\frac{p_\alpha(i)E_i}{\sum_j p_\alpha(j)E_j}
=
\frac{M_i^\alpha}{\sum_j M_j^\alpha}
=
\pi_i(\alpha).
\label{eq:expected-patient-target-share}
\end{equation}
This correction controls target exposure rather than the fraction of
windows drawn from each patient.

\begin{figure}[H]
    \centering
    \includegraphics[width=\textwidth]{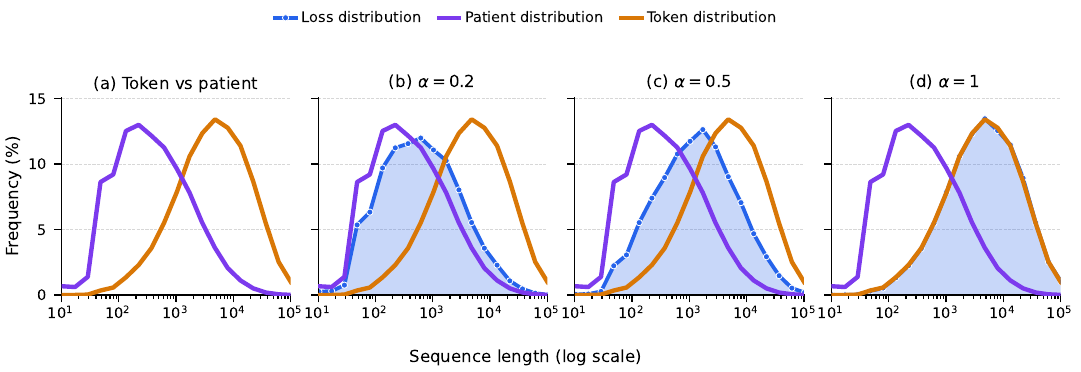}
    \caption{Expected training-target exposure by patient trajectory
    length under Patient Sampling. (a) Distributions of patients and
    tokens. (b--d) Expected target shares for $\alpha=0.2$, $0.5$,
    and $1$. Each distribution sums to 100\% across length bins.}
    \label{fig:loss-share-vs-alpha}
\end{figure}

\subsection{Recursive Patient Sampling}
\label{app:recursive-patient-sampling-weights}

RPS draws chunks with different remaining capacities. When $R$
positions remain, patient $i$ has $T_i+R-3$ possible chunk starts,
and each eligible target occurs in $R-1$ candidate chunks. Before
demographic processing, the expected target count from that draw is
\begin{equation}
E_i(R)=\frac{(R-1)M_i}{T_i+R-3}.
\end{equation}
RPS uses one patient distribution across all capacities. We estimate
how often each capacity occurs by simulating 4,096 packed sequences.
If $c(R)$ is the resulting count, we set
\begin{equation}
\widehat E_i=\sum_{R=2}^{L+1}c(R)E_i(R),
\qquad
p_\alpha(i)=
\frac{M_i^\alpha/\widehat E_i}
{\sum_j M_j^\alpha/\widehat E_j}.
\label{eq:recursive-patient-sampling}
\end{equation}
Starting from the PS weights, we simulate capacities and update these
weights three times. This is an approximate correction because the
capacity distribution depends on the weights and demographic
processing can change chunk lengths.

\subsection{Ordered Global Stream}
\label{app:ordered-global-stream-weights}

OGS distributes each patient's weight across their tokens. If $a(k)$
is the patient at stream position $k$, the token weight is
$q_k=M_{a(k)}^\alpha/T_{a(k)}$. The weights of patient $i$'s tokens
therefore sum to $M_i^\alpha$. For each valid window start $j$, OGS
uses the mean token weight
\begin{equation}
w_j=\frac{1}{L+1}\sum_{k=j}^{j+L}q_k,
\qquad
p_j=\frac{w_j}{\sum_{r\in\mathcal I}w_r}.
\label{eq:ordered-global-stream-weights}
\end{equation}
Sampling windows according to $p_j$ shifts exposure toward the
desired patient shares. The shares are approximate because overlapping
windows can contain different numbers of eligible targets from each
patient, particularly near boundaries.

\subsection{Training Data Coverage}

\begin{figure}[H]
    \centering
    \includegraphics[width=\textwidth]{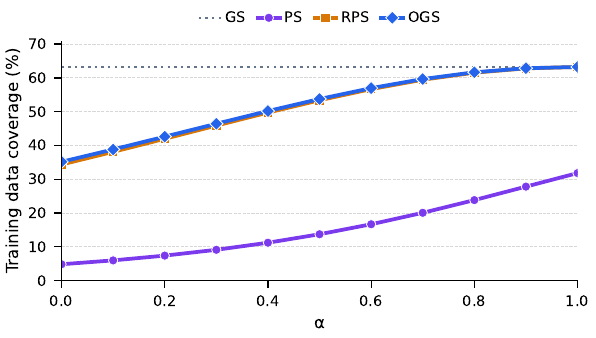}
    \caption{
Expected training-data coverage on MIMIC-IV v2.2 as a function of $\alpha$. Let $u_k$ be the probability that corpus token position $k$ appears in one sampled window. After $B$ independent draws, its probability of being encountered at least once is $1-(1-u_k)^B$; coverage averages this quantity over all corpus positions, regardless of loss eligibility. We use $B\approx141{,}593$ windows with context length 2,048, giving a nominal token budget equal to the training-corpus size. The dotted line shows GS, which is independent of $\alpha$.
    }
    \label{fig:training_data_coverage_vs_alpha}
\end{figure}

\clearpage
\section{Training Recipe}
\label{app:training_recipe}

\begin{table}[H]
\centering
\small
\caption{Model architecture and pretraining configuration. The parameter
count is for MIMIC-IV v2.2 with ETHOS tokenization; it varies with
vocabulary size in the other settings.}
\label{tab:training-recipe}
\begin{tabular}{@{}ll@{}}
\toprule
Parameter & Value \\
\midrule
Architecture & GPT-2-style decoder \citep{GPT2} \\
Parameters & 49,454,592 \\
Transformer layers & 6 \\
Hidden dimension & 768 \\
Attention heads & 12 \\
Feed-forward dimension & 3,072 \\
Positional encoding & Rotary (base 10,000) \\
Context length $L$ & 2,048 \\
Dropout & 0.1 \\
\midrule
Batch size & 128 \\
Optimiser & AdamW \\
AdamW $\beta_1,\beta_2$ & $0.9,\ 0.95$ \\
Weight decay & 0.1 \\
Peak learning rate & $6\times10^{-4}$ \\
Learning-rate warmup & 4,000 steps \\
Learning rate held at peak & 136,000 steps \\
Linear learning-rate decay & 260,000 steps \\
Final learning rate & $6\times10^{-5}$ \\
Training steps & 500,000 \\
Validation interval & 10,000 steps \\
Training precision & BFloat16 autocast \\
\bottomrule
\end{tabular}
\end{table}

The models use untied input and output token embeddings. We train from
scratch on a single GPU for 500,000 optimiser steps, without early
stopping. The learning rate increases linearly during warmup, remains at
its peak for 136,000 steps, then decreases linearly to one tenth of the
peak over 260,000 steps. It remains at that value for the final 100,000
steps. The training loss is the mean cross-entropy over loss-eligible
next-token targets.

For GS and GS+M, every valid global-stream window start is a training
candidate (stride 1); OGS uses candidate starts separated by 64 tokens.
All methods use the \texttt{keep\_left\_most} demographic strategy, which moves all demographic tokens into the leftmost positions of each patient's sequence.
Validation uses fixed windows from the concatenated validation
trajectories with stride 2,048. We evaluate every 10,000 steps and select
the checkpoint with the lowest token-mean validation loss. GS uses
standard causal attention; the other methods restrict attention at
patient or chunk boundaries, as described in Section~3.


\section{Results}
\label{app:results}

\subsection{Initial $\alpha$ Sweep}
\label{app:initial-alpha-sweep}

\begin{figure}[H]
    \centering

    \begin{subfigure}{\textwidth}
        \centering
        \includegraphics[width=0.95\textwidth]
        {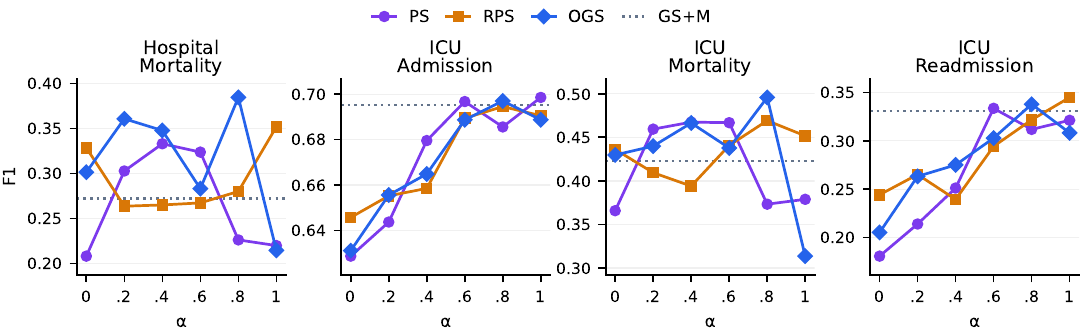}
        \caption{F1 by clinical task.}
        \label{fig:e2_method_sweep_f1}
    \end{subfigure}

    \smallskip

    \begin{subfigure}{\textwidth}
        \centering
        \includegraphics[width=0.95\textwidth]
        {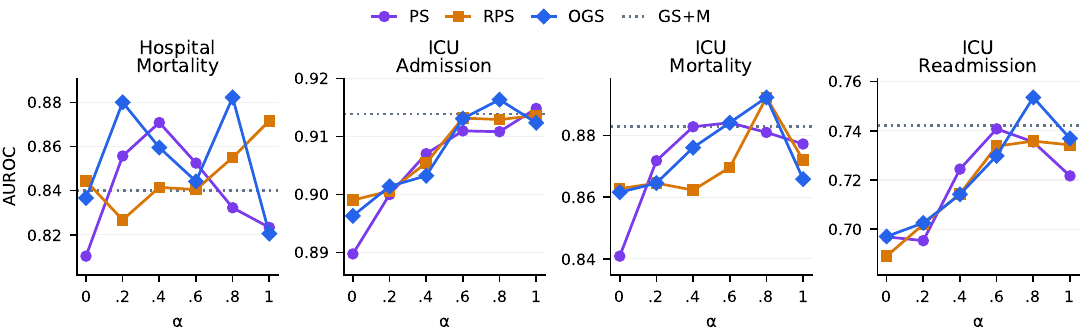}
        \caption{AUROC by clinical task.}
        \label{fig:e2_method_sweep_auroc}
    \end{subfigure}

    \smallskip

    \begin{subfigure}{\textwidth}
        \centering
        \includegraphics[width=0.95\textwidth]
        {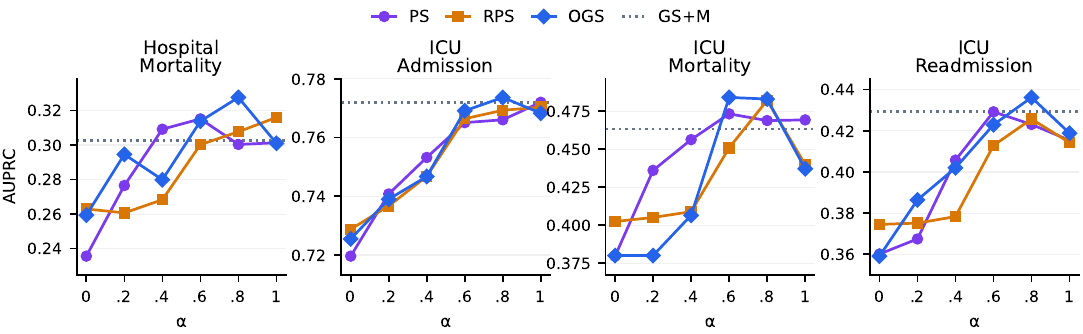}
        \caption{AUPRC by clinical task.}
        \label{fig:e2_method_sweep_auprc}
    \end{subfigure}

    \smallskip

    \begin{subfigure}{\textwidth}
        \centering
        \includegraphics[width=0.95\textwidth]
        {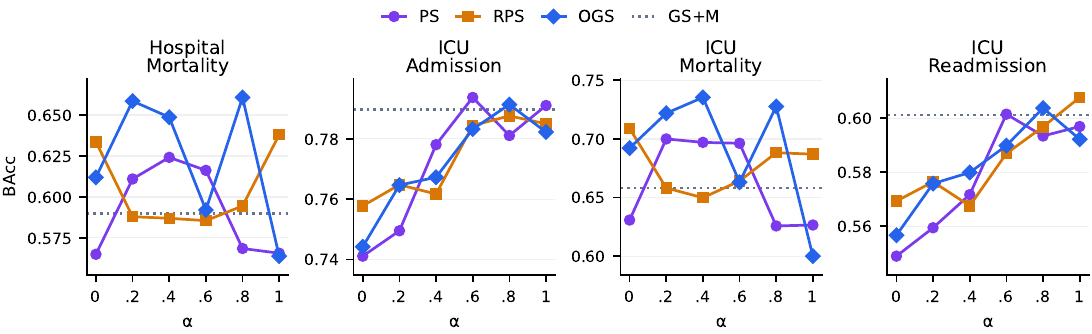}
        \caption{Balanced accuracy by clinical task.}
        \label{fig:e2_method_sweep_bacc}
    \end{subfigure}

    \caption{
        Initial patient-weighting sweep on MIMIC-IV v2.2 with ETHOS
        tokenization (seed 42). Each column shows one clinical prediction
        task; rows show F1, AUROC, AUPRC, and balanced accuracy.
        Curves compare PS, RPS, and OGS across values of the sampling
        exponent $\alpha$. Dotted lines show the GS+M baseline, which
        does not depend on $\alpha$.
    }
    \label{fig:e2_method_sweep}
\end{figure}

\begin{table*}
\centering
\small
\setlength{\tabcolsep}{6pt}
\renewcommand{\arraystretch}{1.05}

\caption{
Initial single-seed patient-weighting sweep on MIMIC-IV v2.2
(seed 42). Downstream macro metrics average four clinical tasks.
Best validation loss is the lowest loss reached during each run.
Bold indicates the best value within each method for each metric.
}
\label{tab:initial-alpha-sweep}

\begin{tabular}{@{}lcccccc@{}}
\toprule
Method
& $\alpha$
& \shortstack{Best validation\\loss $\downarrow$}
& \shortstack{Macro\\AUROC $\uparrow$}
& \shortstack{Macro\\AUPRC $\uparrow$}
& \shortstack{Macro\\F1 $\uparrow$}
& \shortstack{Macro\\BAcc $\uparrow$} \\
\midrule

\multirow{6}{*}{PS}
& $0.0$ & $0.9310$          & $0.8095$          & $0.4238$          & $0.3459$          & $0.6215$ \\
& $0.2$ & $0.9048$          & $0.8307$          & $0.4552$          & $0.4050$          & $0.6550$ \\
& $0.4$ & $0.8826$          & $0.8463$          & $0.4811$          & $0.4328$          & $0.6677$ \\
& $0.6$ & $0.8673$          & $\mathbf{0.8471}$ & $\mathbf{0.4957}$ & $\mathbf{0.4553}$ & $\mathbf{0.6769}$ \\
& $0.8$ & $0.8588$          & $0.8399$          & $0.4896$          & $0.3992$          & $0.6422$ \\
& $1.0$ & $\mathbf{0.8556}$ & $0.8343$          & $0.4895$          & $0.4046$          & $0.6451$ \\
\midrule

\multirow{6}{*}{RPS}
& $0.0$ & $0.9305$          & $0.8238$          & $0.4422$          & $0.4137$          & $0.6674$ \\
& $0.2$ & $0.9103$          & $0.8235$          & $0.4444$          & $0.3984$          & $0.6468$ \\
& $0.4$ & $0.8908$          & $0.8309$          & $0.4506$          & $0.3893$          & $0.6415$ \\
& $0.6$ & $0.8719$          & $0.8393$          & $0.4826$          & $0.4229$          & $0.6553$ \\
& $0.8$ & $0.8595$          & $\mathbf{0.8490}$ & $\mathbf{0.4962}$ & $0.4414$          & $0.6668$ \\
& $1.0$ & $\mathbf{0.8576}$ & $0.8479$          & $0.4851$          & $\mathbf{0.4598}$ & $\mathbf{0.6794}$ \\
\midrule

\multirow{6}{*}{OGS}
& $0.0$ & $0.9311$          & $0.8229$          & $0.4310$          & $0.3919$          & $0.6513$ \\
& $0.2$ & $0.9098$          & $0.8372$          & $0.4500$          & $0.4299$          & $0.6802$ \\
& $0.4$ & $0.8905$          & $0.8383$          & $0.4588$          & $0.4386$          & $0.6828$ \\
& $0.6$ & $0.8722$          & $0.8428$          & $0.4975$          & $0.4281$          & $0.6571$ \\
& $0.8$ & $0.8609$          & $\mathbf{0.8611}$ & $\mathbf{0.5051}$ & $\mathbf{0.4788}$ & $\mathbf{0.6959}$ \\
& $1.0$ & $\mathbf{0.8582}$ & $0.8339$          & $0.4813$          & $0.3813$          & $0.6346$ \\

\bottomrule
\end{tabular}
\end{table*}

\FloatBarrier
\subsection{Local $\alpha$ Sweep}
\label{app:local-alpha-sweep}

\begin{figure}[H]
    \centering

    \begin{subfigure}{\textwidth}
        \centering
        \includegraphics[width=\textwidth]
        {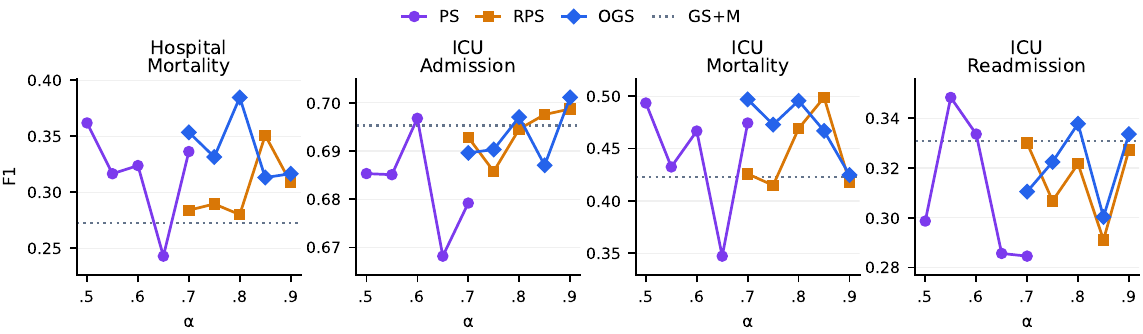}
        \caption{F1 by clinical task.}
        \label{fig:e3_local_alpha_f1}
    \end{subfigure}

    \medskip

    \begin{subfigure}{\textwidth}
        \centering
        \includegraphics[width=\textwidth]
        {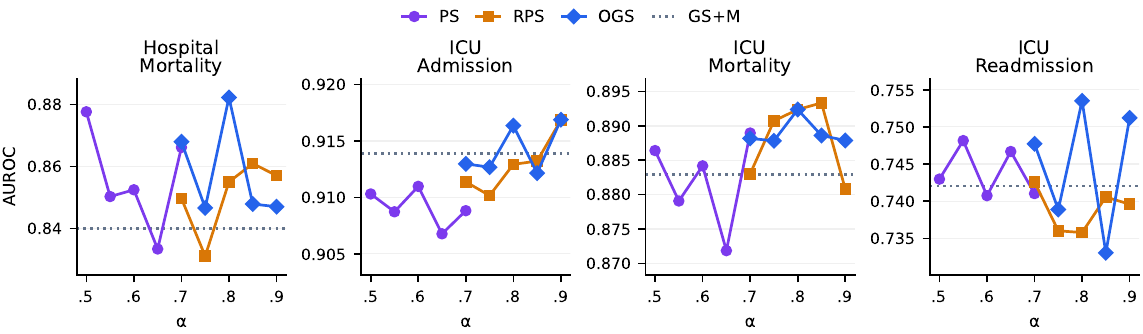}
        \caption{AUROC by clinical task.}
        \label{fig:e3_local_alpha_auroc}
    \end{subfigure}

    \medskip

    \begin{subfigure}{\textwidth}
        \centering
        \includegraphics[width=\textwidth]
        {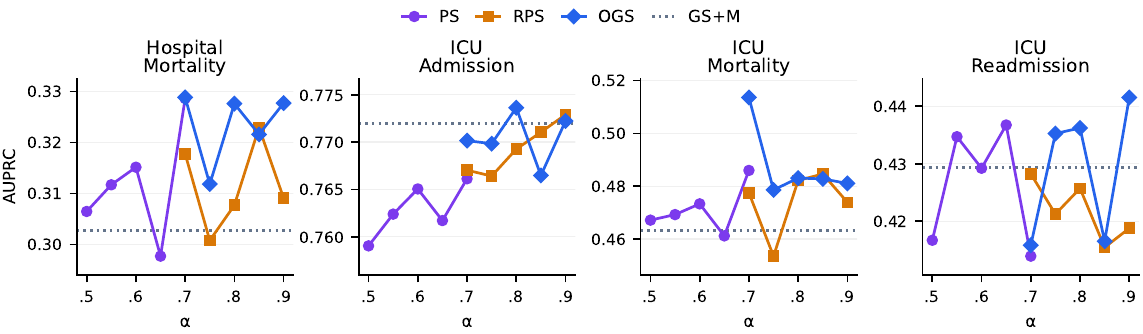}
        \caption{AUPRC by clinical task.}
        \label{fig:e3_local_alpha_auprc}
    \end{subfigure}

    \medskip

    \begin{subfigure}{\textwidth}
        \centering
        \includegraphics[width=\textwidth]
        {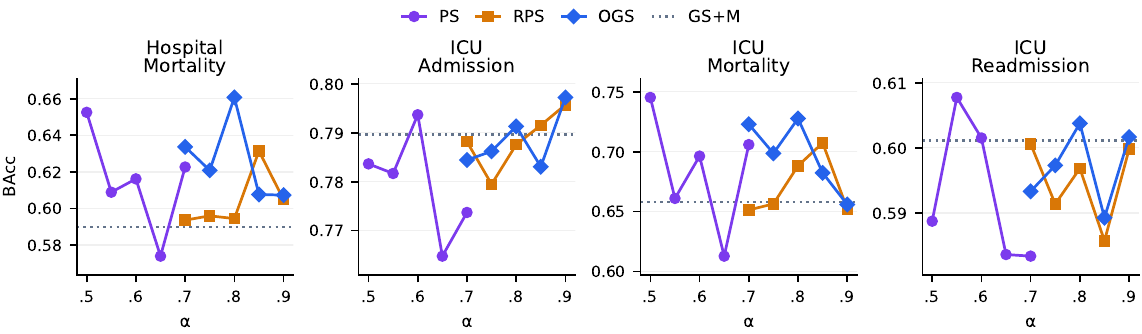}
        \caption{Balanced accuracy by clinical task.}
        \label{fig:e3_local_alpha_bacc}
    \end{subfigure}

    \caption{
        Local patient-weighting sweep on MIMIC-IV v2.2 with ETHOS
        tokenization (seed 42), following the initial coarse sweep.
        Each column shows one clinical prediction task; rows show F1,
        AUROC, AUPRC, and balanced accuracy. Curves compare PS, RPS,
        and OGS across local values of the sampling exponent $\alpha$.
        Dotted lines show the GS+M baseline, which does not depend
        on $\alpha$.
    }
    \label{fig:e3_local_alpha}
\end{figure}

\begin{table*}
\centering
\small
\setlength{\tabcolsep}{6pt}
\renewcommand{\arraystretch}{1.05}

\caption{
Local patient-weighting search on MIMIC-IV v2.2 following the initial
coarse sweep (seed 42). Downstream macro metrics average four clinical
tasks. Best validation loss is the lowest loss reached during each run.
Bold indicates the best value within each method for each metric.
}
\label{tab:local-alpha-search}

\begin{tabular}{@{}lcccccc@{}}
\toprule
Method
& $\alpha$
& \shortstack{Best validation\\loss $\downarrow$}
& \shortstack{Macro\\AUROC $\uparrow$}
& \shortstack{Macro\\AUPRC $\uparrow$}
& \shortstack{Macro\\F1 $\uparrow$}
& \shortstack{Macro\\BAcc $\uparrow$} \\
\midrule

\multirow{5}{*}{PS}
& $0.50$ & $0.8733$          & $\mathbf{0.8543}$ & $0.4874$          & $\mathbf{0.4599}$ & $\mathbf{0.6926}$ \\
& $0.55$ & $0.8703$          & $0.8466$          & $0.4945$          & $0.4456$          & $0.6648$ \\
& $0.60$ & $0.8673$          & $0.8471$          & $0.4957$          & $0.4553$          & $0.6769$ \\
& $0.65$ & $0.8641$          & $0.8397$          & $0.4893$          & $0.3860$          & $0.6338$ \\
& $0.70$ & $\mathbf{0.8623}$ & $0.8512$          & $\mathbf{0.4987}$ & $0.4436$          & $0.6714$ \\
\midrule

\multirow{5}{*}{RPS}
& $0.70$ & $0.8641$          & $0.8467$          & $0.4976$          & $0.4332$          & $0.6584$ \\
& $0.75$ & $0.8617$          & $0.8420$          & $0.4855$          & $0.4243$          & $0.6558$ \\
& $0.80$ & $0.8595$          & $0.8490$          & $0.4962$          & $0.4414$          & $0.6668$ \\
& $0.85$ & $0.8585$          & $\mathbf{0.8520}$ & $\mathbf{0.4985}$ & $\mathbf{0.4596}$ & $\mathbf{0.6790}$ \\
& $0.90$ & $\mathbf{0.8578}$ & $0.8486$          & $0.4937$          & $0.4382$          & $0.6632$ \\
\midrule

\multirow{5}{*}{OGS}
& $0.70$ & $0.8651$          & $0.8542$          & $\mathbf{0.5071}$ & $0.4627$          & $0.6836$ \\
& $0.75$ & $0.8628$          & $0.8465$          & $0.4989$          & $0.4543$          & $0.6757$ \\
& $0.80$ & $0.8609$          & $\mathbf{0.8611}$ & $0.5051$          & $\mathbf{0.4788}$ & $\mathbf{0.6959}$ \\
& $0.85$ & $0.8594$          & $0.8454$          & $0.4968$          & $0.4419$          & $0.6656$ \\
& $0.90$ & $\mathbf{0.8586}$ & $0.8507$          & $0.5056$          & $0.4440$          & $0.6655$ \\
\bottomrule
\end{tabular}
\end{table*}

\FloatBarrier
\subsection{Results}

\begin{table}[H]
\centering
\small
\setlength{\tabcolsep}{8pt}
\renewcommand{\arraystretch}{1.05}

\caption{
Macro downstream performance across four clinical prediction tasks on
MIMIC-IV v2.2, reported as mean $\pm$ standard deviation over five seeds.
GS denotes Global Stream, GS+M Global Stream with masking, PS Patient
Sampling, RPS Recursive Patient Sampling, and OGS Ordered Global Stream.
Best mean results are shown in bold and second-best results in italics.
Higher values are better.
}
\label{tab:main-results}

\begin{tabular}{@{}lccccc@{}}
\toprule
Method
& $\alpha$
& \shortstack{Macro\\F1}
& \shortstack{Macro\\AUROC}
& \shortstack{Macro\\AUPRC}
& \shortstack{Macro\\BAcc} \\
\midrule

GS
& --
& $0.420 \pm 0.017$
& $0.844 \pm 0.004$
& $0.486 \pm 0.010$
& $0.655 \pm 0.010$ \\

GS+M
& --
& $0.430 \pm 0.015$
& $0.845 \pm 0.004$
& $\mathit{0.492 \pm 0.004}$
& $0.660 \pm 0.009$ \\

\cmidrule(lr){1-6}

PS
& $0.70$
& $0.419 \pm 0.032$
& $0.839 \pm 0.012$
& $0.481 \pm 0.018$
& $0.654 \pm 0.018$ \\

RPS
& $0.85$
& $\mathit{0.439 \pm 0.014}$
& $\mathit{0.847 \pm 0.005}$
& $0.492 \pm 0.007$
& $\mathit{0.665 \pm 0.010}$ \\

OGS
& $0.80$
& $\mathbf{0.459 \pm 0.017}$
& $\mathbf{0.849 \pm 0.008}$
& $\mathbf{0.499 \pm 0.007}$
& $\mathbf{0.679 \pm 0.014}$ \\

\bottomrule
\end{tabular}
\end{table}

\clearpage
\section{Tokenization Details}
\label{app:tokenization}

\paragraph{ETHOS tokenization.}
For the MIMIC-IV experiments using ETHOS tokenization, we apply the released ETHOS preprocessing and tokenizer to the hospital, ICU, and emergency department records. ETHOS represents clinical events, demographics, measurements, and elapsed time as discrete tokens. It decomposes hierarchical codes according to their coding systems: for example, ICD-10-CM diagnoses are split into a three-character prefix and subsequent components, while ATC medication codes are split into a prefix, fourth character, and suffix. Eligible numeric measurements are assigned to one of ten quantile bins, with boundaries estimated from the training data. Discrete elapsed-time tokens are inserted between events according to the time interval represented by each gap.

\paragraph{Unified data extraction.}
To test whether the sequence-composition results extend to another EHR representation, we developed dataset-specific DuckDB SQL adapters that convert MIMIC-IV v2.2, eICU-CRD v2.0, and INSPIRE v1.4.2 into chronologically ordered Medical Event Data Standard (MEDS) events. The MIMIC-IV adapter combines hospital and ICU records with linked emergency department records. It includes additional medication administration details, medication reconciliation, microbiology, outpatient measurements, and ICU input, output, and procedure events (Table~\ref{tab:mimic-table-comparison}). The eICU-CRD adapter processes admission information, diagnoses, laboratory results, vital signs, medications and infusions, intake and output, treatments, and respiratory charting. The INSPIRE adapter processes operations, diagnoses, laboratory results, ward and perioperative vital signs, and medications. Where a dataset records times as offsets rather than calendar timestamps, its adapter places events on a consistent synthetic timeline while preserving their relative order and intervals.

We retain patients with at least four distinct event timestamps and split each dataset by patient into training, validation, and test sets. For MIMIC-IV, we map ICD-9 diagnoses and procedures to ICD-10 where possible and link medication identifiers to ATC codes using NDC and GSN mappings. Codes without a suitable mapping retain their source identity where supported.

\paragraph{Shared event tokenization.}
We apply a shared tokenization configuration to the MEDS events from all three datasets, fitting its vocabulary separately on each training split. Each event is serialized into semantic base tokens describing its clinical role and code, followed where applicable by its quantile-binned measurement, unit, or categorical attributes. As in ETHOS, eligible numeric measurements use ten quantile bins. We also use the same elapsed-time tokens and interval rules as ETHOS, inserting them between events at different timestamps. Quantile boundaries and other frequency-dependent tokenization rules are fitted on the training split and applied to the corresponding validation and test splits.

Our representation of hierarchical codes differs from ETHOS. We split a code into its clinical role, coding system, and individual alphanumeric characters. For example, an ICD-10-CM diagnosis of \texttt{I25.10} is represented by the base-token sequence \texttt{DIAGNOSES}, \texttt{ICD-10}, \texttt{I}, \texttt{2}, \texttt{5}, \texttt{1}, \texttt{0}. Medication orders and administrations retain distinct event roles, with dose and route information included when available.

\paragraph{Event-scoped BPE.}
For each dataset, we train a Byte-Pair Encoding (BPE) tokenizer with a budget of up to 1,000 merges. Unlike text BPE, which begins with characters or bytes, our procedure begins with the semantic base tokens described above; each base token is treated as an indivisible symbol. A merge is permitted only within a MEDS event or a group of related event components, such as those describing one medication order. Merges therefore cannot combine unrelated clinical events or cross timestamps. Frequent combinations of codes and their attributes can become compact tokens while the boundaries between successive events remain intact.

\begin{table}[htbp]
\centering
\small
\caption{Selected MIMIC-IV source tables included in the ETHOS and unified tokenization configurations. The unified column reflects the repository's current default MIMIC-IV configuration.}
\label{tab:mimic-table-comparison}
\begin{tabular}{lcc}
\toprule
MIMIC-IV table & ETHOS & Unified \\
\midrule
\multicolumn{3}{l}{\textbf{Demographics and encounters}} \\
\texttt{patients} & \checkmark & \checkmark \\
\texttt{admissions} & \checkmark & \checkmark \\
\texttt{transfers} & \checkmark & \checkmark \\
\texttt{triage} & \checkmark & \checkmark \\
\midrule
\multicolumn{3}{l}{\textbf{Diagnoses and billing procedures}} \\
\texttt{diagnoses\_icd} & \checkmark & \checkmark \\
\texttt{procedures\_icd} & \checkmark & \checkmark \\
\texttt{drgcodes} & \checkmark & \checkmark \\
\texttt{hcpcsevents} & \checkmark & \checkmark \\
\midrule
\multicolumn{3}{l}{\textbf{Medications}} \\
\texttt{prescriptions} & \checkmark & \checkmark \\
\texttt{emar} & $\times$ & \checkmark \\
\texttt{emar\_detail} & $\times$ & \checkmark \\
\texttt{pharmacy} & $\times$ & $\times$ \\
\texttt{medrecon} & $\times$ & \checkmark \\
\texttt{pyxis} & $\times$ & \checkmark \\
\midrule
\multicolumn{3}{l}{\textbf{Laboratory results and vital signs}} \\
\texttt{labevents} & \checkmark & \checkmark \\
\texttt{vitalsign} & \checkmark & \checkmark \\
\texttt{chartevents} & \checkmark & $\times$ \\
\texttt{microbiologyevents} & $\times$ & \checkmark \\
\texttt{omr} & $\times$ & \checkmark \\
\midrule
\multicolumn{3}{l}{\textbf{ICU events}} \\
\texttt{inputevents} & $\times$ & \checkmark \\
\texttt{outputevents} & $\times$ & \checkmark \\
\texttt{procedureevents} & $\times$ & \checkmark \\
\bottomrule
\end{tabular}
\end{table}
\FloatBarrier

\clearpage
\section{Evaluation}
\label{app:evaluation}

\paragraph{Prediction instances.}
We evaluate on the patient-held-out test split. Each occurrence of a task's prediction-point token can produce a separate prediction instance, so a patient may contribute more than one instance. The prompt includes the prediction point and other events recorded at the same timestamp. We retain the demographic prefix and the most recent clinical history within the model's 2,048-token context window. For tasks without a fixed time horizon, the observed label is determined by the first subsequent terminating outcome. Instances with no forward prediction interval are excluded.

The four MIMIC-IV tasks are evaluated with both tokenizers on v2.2 and with ETHOS tokenization on v3.1. INSPIRE supports hospital readmission, ICU admission, and postoperative ICU admission. eICU-CRD supports hospital mortality, ICU readmission, and short ICU stay. Table~\ref{tab:benchmark_definitions} gives each task's prediction point and outcome conditions.

\paragraph{Time-limited tasks.}
Hospital readmission is positive when another hospital admission occurs within 30 days of discharge. Short ICU stay is positive when ICU discharge occurs within 24 hours of admission. If the observed trajectory extends beyond the horizon without the positive outcome, the instance is negative. If it ends before the horizon without a definitive outcome, the instance is excluded because follow-up is incomplete. Observed labels use event timestamps. During generation, elapsed time is estimated from the midpoint durations of the generated ETHOS-style time-interval tokens.

\paragraph{Rollout-based risk estimation.}
For each prediction instance, we sample 20 continuations at temperature 1. Generation stops when a task-specific outcome token appears, when an applicable time horizon is crossed, or when the maximum number of generated tokens is reached. Crossing a time horizon without the positive outcome counts as a negative rollout. Rollouts that reach the token limit without a valid outcome are excluded from the risk estimate. We estimate risk as the fraction of completed rollouts reaching the positive outcome.

The generation limit is 8,192 tokens for all four MIMIC-IV v2.2 tasks with unified tokenization and all three eICU-CRD tasks, for both GS+M and OGS. All remaining settings use 4,096 tokens. Table~\ref{tab:rollout_limits} summarizes these limits.

\paragraph{Metrics and repetitions.}
We calculate AUROC and AUPRC from the prediction-instance risk estimates. F1 and balanced accuracy use a risk threshold of 0.5. Dataset-level macro scores are unweighted means across the tasks evaluated in each dataset and tokenization setting. The primary MIMIC-IV v2.2 comparison reports mean and standard deviation across five independently trained model seeds. Each checkpoint undergoes one stochastic rollout evaluation; we do not separately estimate variation from repeating rollouts for a fixed checkpoint. The four additional dataset and tokenization settings use one trained seed per method.

\begin{table}[htbp]
\centering
\small
\setlength{\tabcolsep}{5pt}
\caption{Prediction points and outcomes for the rollout benchmarks. The positive outcome must precede the competing negative outcome and, where specified, occur within the time horizon. For horizon-based tasks, \texttt{TIMELINE\_END} also stops generated rollouts; observed instances ending before the horizon without a definitive outcome are excluded.}
\label{tab:benchmark_definitions}
\resizebox{\textwidth}{!}{%
\begin{tabular}{@{}llll@{}}
\toprule
Benchmark & Prediction point & Positive / negative outcome & Horizon \\
\midrule
Hospital mortality
& \texttt{HOSPITAL\_ADMISSION}
& \shortstack[l]{\texttt{MEDS\_DEATH} (+)\\\texttt{HOSPITAL\_DISCHARGE} ($-$)}
& -- \\
ICU mortality
& \texttt{ICU\_ADMISSION}
& \shortstack[l]{\texttt{MEDS\_DEATH} (+)\\\texttt{ICU\_DISCHARGE} ($-$)}
& -- \\
ICU readmission
& \texttt{ICU\_DISCHARGE}
& \shortstack[l]{\texttt{ICU\_ADMISSION} (+)\\\texttt{HOSPITAL\_DISCHARGE} ($-$)}
& -- \\
ICU admission
& \texttt{HOSPITAL\_ADMISSION}
& \shortstack[l]{\texttt{ICU\_ADMISSION} (+)\\\texttt{HOSPITAL\_DISCHARGE} ($-$)}
& -- \\
Hospital readmission
& \texttt{HOSPITAL\_DISCHARGE}
& \shortstack[l]{\texttt{HOSPITAL\_ADMISSION} (+)\\\texttt{MEDS\_DEATH} or horizon ($-$)}
& 30 days \\
Postoperative ICU admission
& \texttt{OPERATION\_END}
& \shortstack[l]{\texttt{ICU\_ADMISSION} (+)\\\texttt{HOSPITAL\_DISCHARGE} ($-$)}
& -- \\
Short ICU stay
& \texttt{ICU\_ADMISSION}
& \shortstack[l]{\texttt{ICU\_DISCHARGE} (+)\\\texttt{MEDS\_DEATH} or horizon ($-$)}
& 24 hours \\
\bottomrule
\end{tabular}%
}
\end{table}

\begin{table}[htbp]
\centering
\small
\caption{Maximum generated tokens per rollout. Methods compared within each dataset and tokenization setting use the same limit.}
\label{tab:rollout_limits}
\begin{tabular}{@{}llr@{}}
\toprule
Dataset & Tokenization & Maximum tokens \\
\midrule
MIMIC-IV v2.2 & ETHOS & 4,096 \\
MIMIC-IV v2.2 & Unified & 8,192 \\
MIMIC-IV v3.1 & ETHOS & 4,096 \\
INSPIRE & Unified & 4,096 \\
eICU-CRD & Unified & 8,192 \\
\bottomrule
\end{tabular}
\end{table}

\begin{table}[!htbp]
\centering
\small
\caption{Test-set prediction instances for the indicated dataset and
tokenization settings. A patient may contribute multiple instances.
Prevalence is the percentage of instances with a positive label.}
\label{tab:benchmark-cohorts}
\begin{tabular}{@{}lrrr@{}}
\toprule
Task & Instances & Positive & Prevalence \\
\midrule
\multicolumn{4}{@{}l}{\textbf{MIMIC-IV v2.2, ETHOS tokenization}} \\
Hospital mortality & 42,973 & 836   & 1.95\%  \\
ICU admission      & 42,974 & 6,636 & 15.44\% \\
ICU mortality      & 7,365  & 532   & 7.22\%  \\
ICU readmission    & 6,469  & 850   & 13.14\% \\
\midrule
\multicolumn{4}{@{}l}{\textbf{MIMIC-IV v2.2, unified tokenization}} \\
Hospital mortality & 42,737 & 782 & 1.83\% \\
ICU admission      & 42,737 & 6,557 & 15.34\% \\
ICU mortality      & 7,240  & 531 & 7.33\% \\
ICU readmission    & 6,316  & 792 & 12.54\% \\
\midrule
\multicolumn{4}{@{}l}{\textbf{MIMIC-IV v3.1, ETHOS tokenization}} \\
Hospital mortality & 55,528 & 1,181 & 2.13\% \\
ICU admission      & 55,525 & 8,503 & 15.31\% \\
ICU mortality      & 9,551  & 755 & 7.90\% \\
ICU readmission    & 8,249  & 1,142 & 13.84\% \\
\midrule
\multicolumn{4}{@{}l}{\textbf{INSPIRE, unified tokenization}} \\
Hospital readmission (30 days) & 10,947 & 462 & 4.22\% \\
ICU admission                  & 12,654 & 1,340 & 10.59\% \\
Postoperative ICU admission    & 13,035 & 1,246 & 9.56\% \\
\midrule
\multicolumn{4}{@{}l}{\textbf{eICU-CRD, unified tokenization}} \\
Hospital mortality & 15,731 & 1,455 & 9.25\% \\
ICU readmission    & 17,281 & 2,112 & 12.22\% \\
Short ICU stay (24 hours) & 19,902 & 7,004 & 35.19\% \\
\bottomrule
\end{tabular}
\end{table}
\FloatBarrier

\pagebreak
\FloatBarrier
\section{Dataset Generalisation}
\begin{table}[H]
\centering
\small
\setlength{\tabcolsep}{3pt}
\renewcommand{\arraystretch}{1.05}

\caption{
Task-level generalization results across datasets and tokenizations for seed 42.
GS+M and OGS are compared on each task; BAcc denotes balanced accuracy.
Dataset macro scores are unweighted means over the tasks displayed within each
dataset. The overall mean weights the five dataset settings equally, and OGS
gain is the difference in percentage points, calculated from full-precision
macro scores. Bold indicates the better displayed result in each pair; values
tied after rounding to three decimal places are both bolded. Task definitions
may differ across datasets, so absolute scores should not be compared directly
between them.
}
\label{tab:generalization}

\resizebox{\textwidth}{!}{%
\begin{tabular}{@{}lcccccccc@{}}
\toprule
& \multicolumn{2}{c}{AUROC}
& \multicolumn{2}{c}{AUPRC}
& \multicolumn{2}{c}{F1}
& \multicolumn{2}{c}{BAcc} \\
\cmidrule(lr){2-3}\cmidrule(lr){4-5}
\cmidrule(lr){6-7}\cmidrule(lr){8-9}
Task & GS+M & OGS & GS+M & OGS & GS+M & OGS & GS+M & OGS \\
\midrule

\multicolumn{9}{@{}l}{\textbf{MIMIC-IV v2.2} \hfill \textit{ETHOS tokenization}} \\
Hospital mortality
& $0.836$ & $\mathbf{0.868}$ & $0.310$ & $\mathbf{0.329}$
& $0.265$ & $\mathbf{0.353}$ & $0.585$ & $\mathbf{0.634}$ \\
ICU admission
& $0.911$ & $\mathbf{0.913}$ & $0.769$ & $\mathbf{0.770}$
& $\mathbf{0.690}$ & $\mathbf{0.690}$ & $\mathbf{0.784}$ & $\mathbf{0.784}$ \\
ICU mortality
& $0.886$ & $\mathbf{0.888}$ & $0.462$ & $\mathbf{0.514}$
& $0.431$ & $\mathbf{0.497}$ & $0.662$ & $\mathbf{0.723}$ \\
ICU readmission
& $0.735$ & $\mathbf{0.748}$ & $\mathbf{0.420}$ & $0.416$
& $\mathbf{0.312}$ & $0.311$ & $\mathbf{0.593}$ & $\mathbf{0.593}$ \\
\cmidrule(lr){1-9}
\textit{Macro}
& $0.842$ & $\mathbf{0.854}$ & $0.491$ & $\mathbf{0.507}$
& $0.424$ & $\mathbf{0.463}$ & $0.656$ & $\mathbf{0.684}$ \\
\midrule

\multicolumn{9}{@{}l}{\textbf{MIMIC-IV v2.2} \hfill \textit{Our tokenization}} \\
Hospital mortality
& $0.739$ & $\mathbf{0.831}$ & $0.217$ & $\mathbf{0.267}$
& $0.075$ & $\mathbf{0.185}$ & $0.520$ & $\mathbf{0.555}$ \\
ICU admission
& $0.914$ & $\mathbf{0.916}$ & $0.765$ & $\mathbf{0.766}$
& $0.678$ & $\mathbf{0.681}$ & $0.773$ & $\mathbf{0.776}$ \\
ICU mortality
& $0.773$ & $\mathbf{0.833}$ & $0.313$ & $\mathbf{0.361}$
& $0.176$ & $\mathbf{0.330}$ & $0.549$ & $\mathbf{0.616}$ \\
ICU readmission
& $0.725$ & $\mathbf{0.751}$ & $0.387$ & $\mathbf{0.403}$
& $\mathbf{0.256}$ & $0.243$ & $\mathbf{0.573}$ & $0.569$ \\
\cmidrule(lr){1-9}
\textit{Macro}
& $0.788$ & $\mathbf{0.833}$ & $0.420$ & $\mathbf{0.449}$
& $0.296$ & $\mathbf{0.360}$ & $0.604$ & $\mathbf{0.629}$ \\
\midrule

\multicolumn{9}{@{}l}{\textbf{MIMIC-IV v3.1} \hfill \textit{ETHOS tokenization}} \\
Hospital mortality
& $\mathbf{0.855}$ & $0.851$ & $0.335$ & $\mathbf{0.349}$
& $0.342$ & $\mathbf{0.385}$ & $0.625$ & $\mathbf{0.651}$ \\
ICU admission
& $0.910$ & $\mathbf{0.912}$ & $\mathbf{0.764}$ & $\mathbf{0.764}$
& $0.678$ & $\mathbf{0.681}$ & $0.777$ & $\mathbf{0.781}$ \\
ICU mortality
& $0.865$ & $\mathbf{0.867}$ & $\mathbf{0.476}$ & $0.469$
& $0.450$ & $\mathbf{0.463}$ & $0.673$ & $\mathbf{0.691}$ \\
ICU readmission
& $0.711$ & $\mathbf{0.726}$ & $0.401$ & $\mathbf{0.410}$
& $0.275$ & $\mathbf{0.296}$ & $0.580$ & $\mathbf{0.587}$ \\
\cmidrule(lr){1-9}
\textit{Macro}
& $0.835$ & $\mathbf{0.839}$ & $0.494$ & $\mathbf{0.498}$
& $0.436$ & $\mathbf{0.456}$ & $0.663$ & $\mathbf{0.677}$ \\
\midrule

\multicolumn{9}{@{}l}{\textbf{INSPIRE} \hfill \textit{Our tokenization}} \\
Hospital readmission
& $0.789$ & $\mathbf{0.793}$ & $0.300$ & $\mathbf{0.310}$
& $0.224$ & $\mathbf{0.240}$ & $0.568$ & $\mathbf{0.573}$ \\
ICU admission
& $0.912$ & $\mathbf{0.920}$ & $0.655$ & $\mathbf{0.659}$
& $0.543$ & $\mathbf{0.599}$ & $0.700$ & $\mathbf{0.741}$ \\
Postoperative ICU admission
& $0.957$ & $\mathbf{0.960}$ & $\mathbf{0.831}$ & $0.820$
& $0.755$ & $\mathbf{0.764}$ & $0.834$ & $\mathbf{0.858}$ \\
\cmidrule(lr){1-9}
\textit{Macro}
& $0.886$ & $\mathbf{0.891}$ & $0.595$ & $\mathbf{0.596}$
& $0.507$ & $\mathbf{0.534}$ & $0.700$ & $\mathbf{0.724}$ \\
\midrule

\multicolumn{9}{@{}l}{\textbf{eICU-CRD} \hfill \textit{Our tokenization}} \\
Hospital mortality
& $0.649$ & $\mathbf{0.688}$ & $0.205$ & $\mathbf{0.230}$
& $0.098$ & $\mathbf{0.132}$ & $0.525$ & $\mathbf{0.534}$ \\
ICU readmission
& $\mathbf{0.779}$ & $0.762$ & $\mathbf{0.436}$ & $0.430$
& $\mathbf{0.411}$ & $0.343$ & $\mathbf{0.644}$ & $0.607$ \\
Short ICU stay
& $0.706$ & $\mathbf{0.715}$ & $0.586$ & $\mathbf{0.597}$
& $0.384$ & $\mathbf{0.465}$ & $0.604$ & $\mathbf{0.630}$ \\
\cmidrule(lr){1-9}
\textit{Macro}
& $0.711$ & $\mathbf{0.722}$ & $0.409$ & $\mathbf{0.419}$
& $0.298$ & $\mathbf{0.313}$ & $\mathbf{0.591}$ & $0.590$ \\
\midrule

\multicolumn{9}{@{}l}{\textbf{Across dataset settings} \hfill
\textit{}} \\
\textit{Mean of dataset macros}
& $0.813$ & $\mathbf{0.828}$
& $0.482$ & $\mathbf{0.494}$
& $0.392$ & $\mathbf{0.425}$
& $0.643$ & $\mathbf{0.661}$ \\
\textit{OGS gain}
& \multicolumn{2}{c}{$+1.52$}
& \multicolumn{2}{c}{$+1.22$}
& \multicolumn{2}{c}{$+3.29$}
& \multicolumn{2}{c}{$+1.80$} \\
\bottomrule
\end{tabular}%
}
\end{table}

\end{document}